\documentclass[english]{article}
\usepackage[T1]{fontenc}
\usepackage[latin9]{inputenc}
\usepackage{babel}
\usepackage{array}
\usepackage{multirow}
\usepackage{graphicx}
\usepackage{url}
\usepackage{wrapfig}
\usepackage[table]{xcolor}
\usepackage[unicode=true]
 {hyperref}

\makeatletter

\providecommand{\tabularnewline}{\\}

\usepackage{hyperref}
\usepackage{url}
\usepackage{booktabs}
\usepackage{amsfonts}
\usepackage{nicefrac}
\usepackage{microtype}
\makeatother

\begin{document}

\title{Submanifold Sparse Convolutional Networks}

\author{Benjamin Graham, Laurens van der Maaten\\Facebook AI Research\\\tt{\{benjamingraham,lvdmaaten\}@fb.com}}
\maketitle
\begin{abstract}
Convolutional network are the de-facto standard for analysing spatio-temporal data such as images, videos, 3D shapes, etc. Whilst some of this data is naturally dense (for instance, photos), many other data sources are inherently sparse. Examples include pen-strokes forming on a piece of paper, or (colored) 3D point clouds that were obtained using a LiDAR scanner or RGB-D camera. Standard ``dense'' implementations of convolutional networks are very inefficient when applied on such sparse data. We introduce a sparse convolutional operation tailored to processing sparse data that differs from prior work on sparse convolutional networks in that it operates strictly on submanifolds, rather than ``dilating'' the observation with every layer in the network. Our empirical analysis of the resulting submanifold sparse convolutional networks shows that they perform on par with state-of-the-art methods whilst requiring substantially less computation.
\end{abstract}

\section{Introduction}

Convolutional networks constitute the state-of-the art method for a wide range of tasks that involve the analysis
of data with spatial and/or temporal structure, such as photographs, videos,
or three-dimensional surface models. While such data frequently comprises a densely filled (2D or 3D) grid, other spatio-temporal datasets are naturally sparse. For instance, handwriting is made up of one-dimensional lines in two-dimensional space, pictures made by RGB-D cameras are three-dimensional point clouds, and OFF models
form two-dimensional surfaces in 3D space. The curse of dimensionality applies, in particular,
on data that lives on grids that have three or more dimensions: the number of points on the grid grows exponentially with its dimensionality. In such scenarios, it becomes increasingly important to exploit data sparsity whenever possible in order to reduce the computational resources needed for data processing. Indeed, exploiting sparsity is paramount when analyzing, for instance, RGB-D videos which are sparsely populated 4D structures.

Traditional convolutional network implementations are optimized for data that lives on densely populated grids, and cannot process
sparse data efficiently. More recently, a number of convolutional network implementations have been presented that are tailored to work efficiently on
sparse data \cite{Vote3Deep,Sparse3D,OctNet}. Mathematically, some of these implementations are identical to a regular convolutional network, but they require fewer computational resources in terms of FLOPs \cite{Vote3Deep} and/or in terms of memory \cite{Sparse3D}.
OctNets \cite{OctNet} slightly modify the convolution operator to produce ``averaged'' hidden states in parts of the grid that are away from regions of
interest.

One of the downsides of prior sparse implementations of convolutional networks is that they ``dilate'' the sparse data in every layer, because they implement a ``full'' convolution. In this work, we show that it is possible to successfully train convolutional networks that keep the same sparsity pattern throughout the layers of the network, without dilating the feature maps. To this end, we explore two novel convolution operators: \emph{sparse convolution} (SC) and \emph{valid sparse convolution} (VSC). In our experiments with recognizing handwritten digits and 3D shapes, networks using SC and VSC achieve state-of-the-art performance whilst reducing the computation and memory requirements by $\sim \! 50\%$.

\section{Motivation}
We define a $d$-dimensional convolutional network as a network that
takes as input that is a $(d+1)$-dimensional tensor: the input tensor contains $d$ spatiotemporal
dimensions (such as length, width, height, time, \emph{etc.}) and one additional feature space
dimension (for instance, RGB color channels, surface normal vectors, \emph{etc.}).
A sparse input corresponds to a $d$-dimensional grid of \emph{sites} that is
associated with a feature vector. We define a site in the input to be \emph{active} if any element in the feature vector is not in its \emph{ground state}, for instance, if it is non-zero\footnote{Note that the ground state does not necessarily have to be zero.}. In many practical problems, thresholding
may be used to eliminate sites at which the feature
vector is within a very small distance from the ground state. Note that even though the input tensor is $(d+1)$-dimensional,
activity is a $d$-dimensional phenomenon: entire planes along the feature dimension
are either active or not.

The hidden layers of a convolutional network are also represented by $d$-dimensional
grids of feature-space vectors. When propagating the input data through the network,
a site in a hidden layer is active if any of the sites
in the layer that it takes as input is active. (Note that when using $3\times3$ convolutions,
each site is connected to $3\times3=9$ sites in the hidden layer below.) Activity in a hidden layer thus follows an inductive definition in which each layer determines the set of active states in the next. In each hidden layer, inactive
sites all have the same feature vector: the one corresponding to the ground state. Note that the ground state in a hidden layer
is often not equal to zero, in particular, when convolutions with a bias term
are used. However, irrespective of the value of the ground state, the ground-state value
only needs to be calculated once per forward pass during training (and only once for all forward passes at test time). This allows for substantial savings in computational and memory requirements; the exact savings depend on the data sparsity and the network depth.

In this paper, we argue that the framework described above is unduly restrictive, in particular, because the convolution
operation has not been modified to accommodate the sparsity
of the input data. If the input data contains a single active site, then
after applying a $3^d$ convolution, there will be $3^{d}$ active
sites. Applying a second convolution of the same size will yield $5^{d}$ active sites,
and so on. This rapid growth of the number of active sites is a poor prospect when implementing
modern convolutional network architectures that comprise tens or even hundreds of convolutions, such as VGG networks, ResNets, and DenseNets \cite{ResNet,DenseNet,journals/corr/SimonyanZ14a}.
Of course, convolutional networks are not often applied to inputs that only have a single active site, but the aforementioned ``dilation'' problems are equally problematic when the input data comprises one-dimensional
curves in spaces with two or more dimensions, or two-dimensional surfaces in three or more dimensions.

To address the problems with dilation of active sites, we propose two slightly different
convolution operations for use in convolutional networks. What the two operations have in common is that they both
ignore the ground state: they replace the ground state with a zero vector to simplify
the convolution operations. The difference between both operations is in how they
handle active sites: instead of automatically making a site active
if any of the inputs to its receptive field is active (thereby dilating the set of active sites), our most efficient convolutional operation only considers the central input. As a result, the output set of active sites exactly mirrors that of the input set. We empirically demonstrate that use of our adapted convolutional operators allows us to build much deeper networks that achieve state-of-the-art results whilst requiring much fewer resources by preserving sparsity.

\begin{figure}
\begin{centering}
\includegraphics[width=0.32\columnwidth]{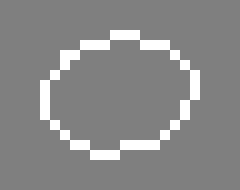}
\includegraphics[width=0.32\columnwidth]{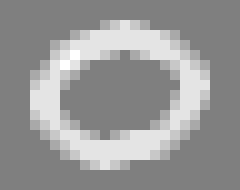}
\includegraphics[width=0.32\columnwidth]{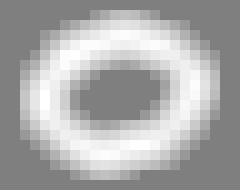}
\caption{Example of ``submanifold'' dilation. \textbf{Left:} Original curve. \textbf{Middle:} Result of applying a regular $3 \times 3$ convolution with weights $1/9$. \textbf{Right:} Result of applying the same convolution again. The example shows that regular convolutions substantially reduce the sparsity of the feature maps.}\label{6pics}
\end{centering}
\end{figure}

\subsection{Submanifold Dilation}
In Figure \ref{6pics}, we show an example of a one-dimensional curve that is embedded
on a two-dimensional grid. The figure shows that even when we apply small $3\times3$ convolutions on this grid, the sparsity on the grid rapidly disappears. At the same time, if we restrict the output of the convolution only to the set of
active input points, hidden layers in the network cannot capture a lot of information that may relevant to the classification of the curve.
In particular, two neighboring connected components will be treated
completely independently. Luckily, nearly all convolutional networks incorporate
some form of pooling, or use strided convolutions. These operations are essential in the sparse convolutional networks\footnote{By ``sparse convolutional networks'', we mean networks designed to operate on sparse input data. We do not mean networks that have sparse parameter matrices \cite{braindamage,liupensky}.} we investigate, as they allow neighboring components to merge. In particular, the closer the components are,
the smaller the number of poolings / strided convolutions is that is necessary for the components to merge in the hidden-layer representations.

\subsection{Very Deep Convolutional Networks}

In image classification, very deep convolutional networks with
small filters, often of size $3\times3$ pixels and a padding of $1$ pixel (to preserve the size of the feature maps), have proven to be very effective. Such small filters were used successfully in VGG networks, which have relatively wide layers. The introduction of residual networks (ResNets) showed that deeper but
narrow networks with small filters are more efficient. The success of very deep ResNets, ResNeXt models, and
DenseNets with bottleneck connections \cite{ResNet,DenseNet,Xie2016} shows that it can be useful
to calculate a relatively small number of features at a time and amalgamate
these features into a larger state variable, either by vector-addition or feature-vector
concatenation.

Unfortunately, these techniques are impractical using existing sparse
convolutional network implementations. One problem is that networks with multiple
paths will tend to generate different sets of active paths, which
would have to be merged to reconnect the outputs. It seems that this
would be difficult to perform this merging efficiently. More importantly, ResNets
and DenseNets generate such large receptive fields that sparsity would
almost immediately be destroyed by the explosion in the number of
active sites.

\section{(Valid) Sparse Convolutions: SC and VSC}

We define a \emph{sparse convolution} SC($m,n,f,s)$ with $m$ input
feature planes, $n$ output feature planes, a filter size of $f$,
and stride $s$. We assume $f$ and $s$ to be odd integers, but
we can allow generalization to non-square filters, \emph{e.g.}, $f=1\times7$
or $f=7\times1$, if we want to implement Inception-style factorised
convolutions \cite{journals/corr/SzegedyLJSRAEVR14}. An SC convolution computes the set of active
sites in the same way as a regular convolution: it looks for the presence of any active sites
in its receptive field of size $f^{d}$. If the input has size $\ell$
then the output will have size $(\ell-f+s)/s$. An SC convolution
differs from a regular convolution in that it discards the ground
state for non-active sites by assuming that the input from those sites
is exactly zero. Whereas this is a seemingly small change to the convolution operation, it may bring
computational benefits in practice.

Next, we define a second type of sparse convolution, which forms
the main contribution of this paper. Again, let $f$ denote an odd number,
or collection of odd numbers, \emph{e.g.}, $f=3$ or $f=1\times7$. We define a \emph{valid
sparse convolution} VSC$(m,n,f,1)$ as a modified SC$(m,n,f,1)$
convolution. First, we pad the input with $(f-1)/2$ on each side,
so that the output will have the same size as the input. Next, we restrict an output
site to be active if and only if the site at the corresponding
site in the input is active (\emph{i.e.}, if the central site in the receptive field is active). Whenever an output site is determined to be active, its output feature vector is calculated by the SC operation. Table \ref{tbl:flops} presents the computational and memory requirements of a regular convolution (C) and of our SC and VSC convolutions.

To construct convolutional networks using SC and VSC, we also need activation functions, batch normalization, and pooling. Activation functions are defined as usual, but are restricted to the set of active sites. Similarly, we define batch normalization in terms of regular batch-normalization applied over the set of active sites. Max-pooling MP$(f,s)$ and average-pooling AP$(f,s)$ operations are defined as a variant of SC$(\cdot,\cdot,f,s)$. MP takes the maximum of the zero
vector and the input feature vectors in the receptive field. AP calculates $f^{-d}$ times the sum of the active input vectors.

We also define a deconvolution operation DC$(\cdot,\cdot,f,s)$ as an inverse of the SC$(\cdot,\cdot,f,s)$ convolution \cite{zeiler10}. The set of active output sites from a DC convolution is exactly the set of input active sites to the matching SC convolution. The set of connections between input-output sites is simply inverted.

\begin{table}
\caption{Computational and memory costs of three different convolutional operations at active and non-active sites: regular convolution (C), sparse convolution (SC), and valid sparse convolution (VSC). We consider convolutions of size 3 at a single location
in $d$ dimensions. Notation: $a$ is the number of active inputs to
the spatial location, $m$ the number of input feature planes, and
$n$ the number of output feature planes.}\label{tbl:flops}
\vspace{3pt}
\centering{}
\begin{tabular}{ll|ccc}\toprule
\textbf{Active} & \textbf{Type} & \textbf{C} & \textbf{SC} & \textbf{VSC}\\\midrule
\multirow{2}{*}{No} &  FLOPs & $3^{d}mn$ & $amn$ & 0\\
 &  Memory & $n$ & $n$ & 0\tabularnewline
\hline
\multirow{2}{*}{Yes} &  FLOPs & $3^{d}mn$ & $amn$ & $amn$\\
 &  Memory & $n$ & $n$ & $n$ \\\bottomrule
\end{tabular}
\end{table}

\subsection{Submanifold Convolutional Networks}

\begin{figure}
\begin{centering}
\par\end{centering}
\includegraphics[width=\linewidth]{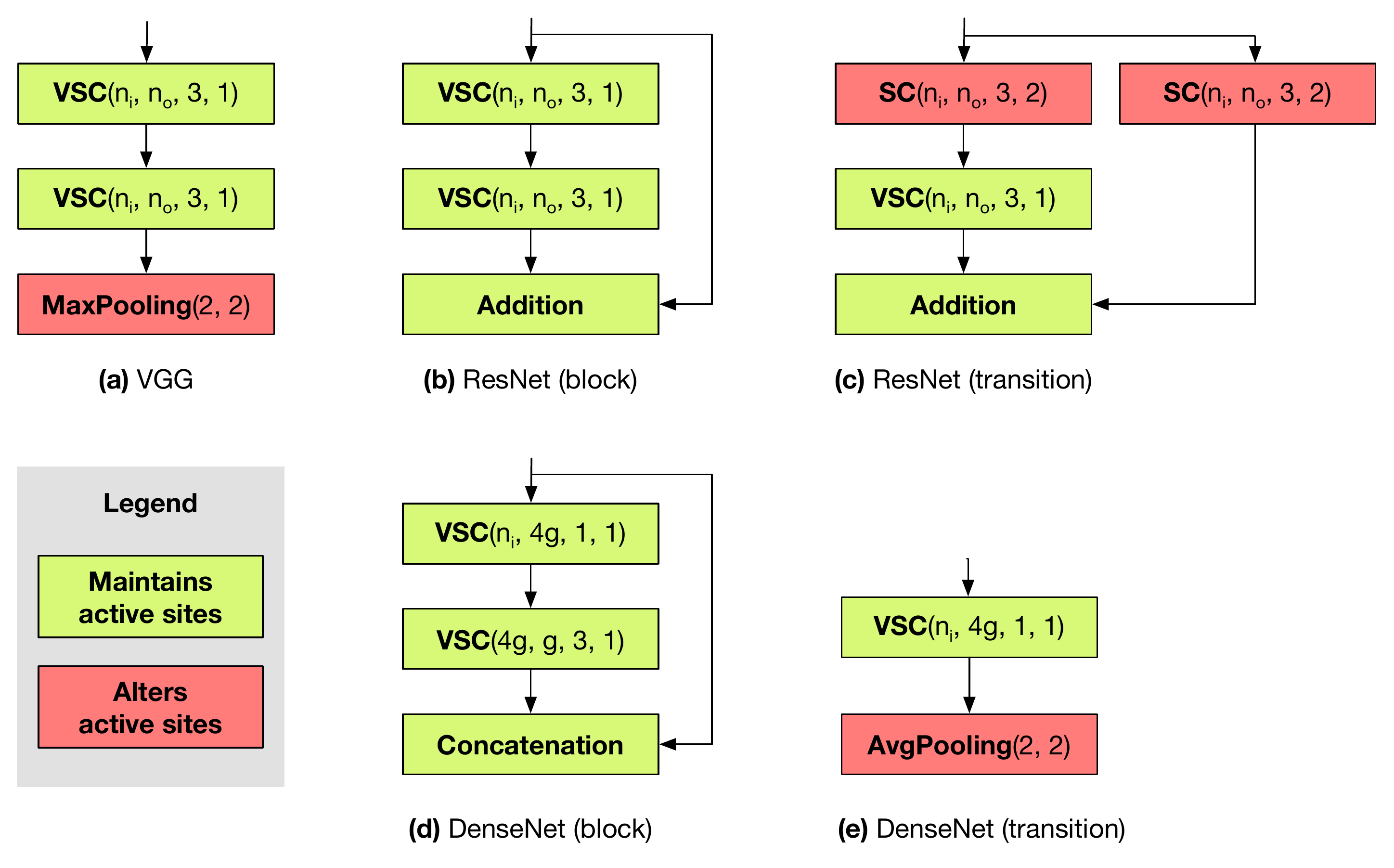}
\caption{Modules used for building sub-manifold sparse convolutional networks: \textbf{(a)} VGG blocks comprise two VSC convolutions and a max-pooling operation; \textbf{(b)} ResNet blocks that maintain spatial resolution add the output of two VSC convolutions to the input; \textbf{(c)} ResNet blocks that reduce spatial resolution replace the first VSC convolution and the (implicit) identity function by a strided SC convolution; \textbf{(d)} DenseNet blocks that maintain spatial resolution concatenate the output of two VSC convolutions with the input; and \textbf{(e)} DenseNet blocks that reduce spatial resolution perform a single VSC convolution and an average-pooling. The four arguments of a convolution operator (SC or VSC) are the number of input planes $n_i$, the number of output planes $n_o$, the kernel size, and the stride, respectively. The two arguments of a pooling operator are the kernel size and the stride, respectively. The ``growth rate'' of a DenseNet \cite{DenseNet} is denoted by $g$.}\label{modules}
\end{figure}

We use a combination of VSC convolutions, strided SC convolutions, and sparse
pooling operations to build sparse versions of the popular VGG,
ResNet, and DenseNet convolutional networks. The blocks we use in our networks are presented in Figure \ref{modules}. We refer to our networks as \emph{submanifold convolutional networks}, because they are optimised to process
low-dimensional data living in a space of higher dimensionality.\footnote{We note that this is a slight abuse of the term ``submanifold''.
Our input data may contain multiple connected components, and even a mixture of 1D and 2D objects embedded
in 3D space.}

We use the name VGG to refer to networks that contain a number
of VSC($\cdot$,$\cdot$,3,1) convolutions, separated by max-pooling \cite{journals/corr/SimonyanZ14a}.
Each convolution is followed by batch normalization and a ReLU non-linearity.

Similarly, we define ``pre-activated ResNets'' \cite{ResNet} in which most data processing is performed by pairs of VSC($\cdot$,$\cdot$,3,1) convolutions, and in which the residual connections are identity functions. Whenever the number
of input / output features is different, we use a VSC($\cdot$,$\cdot$,1,1) instead. Whenever there is change of scale, we replace the first convolution and the residual connection by a SC($\cdot$, $\cdot$,3,2) convolution. This ensures that
two branches can use the same hash table of active sites, and reduces additions
to a simple sum of two equally sized matrices. The
increased size of the residual connection's receptive field also prevents
excessive information loss.

We also experiment with submanifold DenseNets \cite{DenseNet}. Herein, the word \emph{dense} does not refer to a lack of spatial sparsity but rather to the pattern of connections
between convolution operations. A simple DenseNet module is a sequence
of convolutions in which each convolution takes as input the concatenated
output of \emph{all} the previous convolution operations. The bottleneck layers in our submanifold DenseNets are implemented in the same way as for ResNets.

\section{Implementation}

To implement (V)SC convolutions efficiently, we store the
state of a input/hidden layer in two parts: a hash table\footnote{\url{https://github.com/sparsehash/sparsehash}} and a matrix.
The matrix has size $a \times m$ and contains one row for each of the $a$
active sites. The hash table contains (location, row) pairs
for all active sites: the location is a tuple of integer coordinates,
and the row number indicates the corresponding row in the feature
matrix. Given a convolution with filter size $f$, we define a \emph{rule book} to be a collection $R=(R_{i}:i\in\{0,1,...,f-1\}^{d})$
of $f^{d}$ integer matrices of size $k_{i}\times2$. To implement
an SC($m,n,f,s)$ convolution, we:

\begin{itemize}
\item Iterate once through the the input hash-table. We build the output hash table and rule book on-the-fly by iterating over
points in the output layer that receive input from a given point in the input
layer. When an output site is visited for the first time, a new entry
is created in the output hash table. Based on the spatial offset between
the input and output points, a (input index, output index) pair is added
to the rule book.

\item Initialize the output matrix to all zeros. For each $i\in f$, there is a parameter matrix $W^{i}$ with
size $m\times n$. For each $j\in\{1,\dots,k_{i}\}$, multiply the
$R^{i}(j,1)$-th row of the input feature matrix by $W^{i}$ and add
it to the $R^{i}(j,2)$-th row of the the output feature matrix. This
can be implemented very efficiently on GPUs because it is a
matrix-matrix multiply-add operation.
\end{itemize}

To implement a VSC convolution, we re-use the input hash table for the
output, and construct an appropriate rule book. Note that because the sparsity pattern does not change, the same rule book
can be re-used in VGG/ResNet/DenseNet networks until a pooling or subsampling layer is encountered.

If there are $a$ active points in the input layer, the cost
of building the input hash-table is $O(a)$. For VGG/ResNet/DenseNet
networks, assuming the number of active sites reduces by a multiplicative
factor with each pooling operation, the cost of building all the
hash-tables and rule-books is also $O(a)$, regardless of the depth of
the network.

\section{Experiments}
We perform experiments on a 2D and a 3D dataset with sparse images. The \textbf{CASIA dataset} \cite{casia}\footnote{\url{http://www.nlpr.ia.ac.cn/databases/handwriting/Online_database.html}} contains samples of 3755 GBK level-1 characters with approximately
240 train and 60 test images per class. CJVK characters are good test cases for our models because they are a worst-case scenario for sparse convolutional networks: when drawn at scale $64\times64$, about 8\% of the pixels are active, but this percentage rapidly decreases after pooling due to the small density of the pen strokes. This makes them a good test case for our models. The \textbf{ModelNet-40 dataset}\footnote{\url{http://3dshapenets.cs.princeton.edu/}}
contain 2468 CAD models that contain shapes corresponding to 40 classes. We follow the preprocessing
of \cite{oai:arXiv.org:1604.03265} before feeding the models into our convolutional networks. All CAD models were rendered as surfaces at size $30^{3}$.

\subsection{Results on CASIA}
We first experiment with two VGG architectures on the CASIA dataset. We trained all models for 100 epochs using batches of size 100, SGD with momentum 0.9, a weight decay of $10^{-4}$, and a learning
rate decay of 5\% per epoch. For simplicity, we do not employ any data augmentation.

The architectures of our VGG networks and their performances are presented in Table \ref{vgg}. We observe that ``regular'' C convolutions and ``sparse'' SC convolutions achieve the same error: this result suggests that discarding the ground state has essentially no negative impact on performance. This is an argument for always discarding ground states, as it makes things easier computationally and algorithmically. Comparing SC with VSC convolutions, we observe a minimal loss in performance by considering only the valid part of the convolution. This minimal loss in accuracy does facilitate great computational improvements: networks using VSC use 2 to 3$\times$ less computation and memory.

\begin{table}
\begin{centering}
\caption{Classification error, computational requirements (in FLOPs), and memory load (measured by the number of hidden states) of two VGG networks with three different convolutional operations (C, SC, and VSC) on the CASIA dataset. The network architecture is specified in the top part of the table; the bottom part presents the performance characteristics of the networks. Lower is better.\label{vgg}}
\vspace{3pt}
\par\end{centering}
\centering{}%
\begin{tabular}{l|ccc|ccc}
\toprule
\multicolumn{1}{c|}{\textbf{Feature size}} & \multicolumn{3}{c|}{\textbf{VGG-A}} & \multicolumn{3}{c}{\textbf{VGG-B}}\tabularnewline
\midrule
\multicolumn{1}{c|}{64$\times$64}  & \multicolumn{3}{c|}{2$\times$16C3, MP2} & \multicolumn{3}{c}{2$\times$16C3, MP2}\tabularnewline
\multicolumn{1}{c|}{32$\times$32} & \multicolumn{3}{c|}{2$\times$32C3, MP2} & \multicolumn{3}{c}{2$\times$32C3, MP2}\tabularnewline
\multicolumn{1}{c|}{16 $\times$16} & \multicolumn{3}{c|}{2$\times$48C3, MP2} & \multicolumn{3}{c}{2$\times$64C3, MP2}\tabularnewline
\multicolumn{1}{c|}{8$\times$8} & \multicolumn{3}{c|}{2$\times$64C3, MP2} & \multicolumn{3}{c}{2$\times$128C3, MP2}\tabularnewline
\multicolumn{1}{c|}{4$\times$4} & \multicolumn{3}{c|}{2$\times$96C43} & \multicolumn{3}{c}{2$\times$256C3}\tabularnewline
\multicolumn{1}{c|}{4$\times$4} & \multicolumn{3}{c|}{128C4} & \multicolumn{3}{c}{512C4}\tabularnewline
\toprule
\textbf{Convolution} & \textbf{C} & \textbf{SC} & \textbf{VSC} & \textbf{C} & \textbf{SC} & \textbf{VSC}\tabularnewline
\midrule
\textbf{Class. error} (in \%) & 4.15 & 4.11 & 4.67 & 3.45 & 3.47 & 3.82\tabularnewline
\textbf{FLOPs} ($\times10^{6}$) & 41 & 25 & 7.4 & 72 & 50 & 23\tabularnewline
\textbf{Hidden states} ($\times10^{3}$) & 233 & 133 & 41 & 254 & 155 & 55\tabularnewline
\bottomrule
\end{tabular}
\end{table}

Next, we performed experiments on CASIA with submanifold ResNets.
The key difference between our implementation of ResNets and regular
ResNets is that stride-2 ResNet modules use SC($\cdot$,$\cdot$,3,2) convolutions
for the strided convolution, rather than SC($\cdot$,$\cdot$,1,2). This change is necessary
to ensure the two branches produce the same set of active sites, which simplifies
bookkeeping and turns the addition operation into a simple matrix-matrix
addition. Unlike the VSC convolutions that are used in most layers,
the SC($\cdot$,$\cdot$,3,2) we use after downsampling leads sites to be active if \emph{any} of its inputs
are active, which avoids information loss in the transition. The architectures of our ResNet networks and their performances are presented in Table \ref{resnet results}. The results with ResNets are in line with those obtained using VGG networks: we obtain reductions in computational and memory requirements by at least a factor of 2 at a minimal loss in accuracy.

\begin{table}
\caption{Classification error, computational requirements (in FLOPs), and memory load (measured by the number of hidden states) of four ResNet networks with two different convolutional operations (C and VSC) on the CASIA dataset. The network architecture is specified in the top part of the table; the bottom part presents the performance characteristics of the networks. Lower is better.\label{resnet results}}
\vspace{3pt}
\centering{}
\resizebox{\linewidth}{!}
{
\begin{tabular}{l|cc|cc|cc|cc}
\toprule
\multicolumn{1}{c|}{\textbf{Feature size}} & \multicolumn{2}{c|}{\textbf{ResNet-A}} & \multicolumn{2}{c|}{\textbf{ResNet-B}} & \multicolumn{2}{c|}{\textbf{ResNet-C}} & \multicolumn{2}{c}{\textbf{ResNet-D}}\tabularnewline
\midrule
\multicolumn{1}{c|}{64$\times$64} & \multicolumn{2}{c|}{16C3, MP2} & \multicolumn{2}{c|}{16C3, MP2} & \multicolumn{2}{c|}{16C3, MP2} & \multicolumn{2}{c}{16C3, 2$\times$ResNet(16)}\tabularnewline
\multicolumn{1}{c|}{32$\times$32} & \multicolumn{2}{c|}{2$\times$ResNet(16)} & \multicolumn{2}{c|}{2$\times$ResNet(16)} & \multicolumn{2}{c|}{2$\times$ResNet(16)} & \multicolumn{2}{c}{2$\times$ResNet(32)}\tabularnewline
\multicolumn{1}{c|}{16$\times$16} & \multicolumn{2}{c|}{2$\times$ResNet(32)} & \multicolumn{2}{c|}{2$\times$ResNet(32)} & \multicolumn{2}{c|}{2$\times$ResNet(32)} & \multicolumn{2}{c}{2$\times$ResNet(48)}\tabularnewline
\multicolumn{1}{c|}{8$\times$8} & \multicolumn{2}{c|}{2$\times$ResNet(48)} & \multicolumn{2}{c|}{2$\times$ResNet(64)} & \multicolumn{2}{c|}{2$\times$ResNet(64)} & \multicolumn{2}{c}{2$\times$ResNet(64)}\tabularnewline
\multicolumn{1}{c|}{4$\times$4} & \multicolumn{2}{c|}{2$\times$ResNet(96)} & \multicolumn{2}{c|}{2$\times$ResNet(128)} & \multicolumn{2}{c|}{2$\times$ResNet(128)} & \multicolumn{2}{c}{2$\times$ResNet(128)}\tabularnewline
\multicolumn{1}{c|}{4$\times$4} & \multicolumn{2}{c|}{128C4} & \multicolumn{2}{c|}{256C4} & \multicolumn{2}{c|}{2$\times$ResNet(256)} & \multicolumn{2}{c}{2$\times$ResNet(256)}\tabularnewline
\multicolumn{1}{c|}{2$\times$2} & \multicolumn{2}{c|}{--} & \multicolumn{2}{c|}{--} & \multicolumn{2}{c|}{512C2} & \multicolumn{2}{c}{512C2}\tabularnewline
\toprule
\textbf{Convolution} & \textbf{C} & \textbf{VSC} & \textbf{C} &  \textbf{VSC} & \textbf{C} & \textbf{VSC} & \textbf{C} & \textbf{VSC}\tabularnewline
\midrule
\textbf{Class. error} (in \%) & 3.97 & 4.09 & 3.78 & 3.87 & 3.70 & 3.78 & 3.51 & 3.57\tabularnewline
\textbf{FLOPs} ($\times10^{6}$) & 32 & 14.5 & 40 & 21 & 51 & 31 & 88 & 49\tabularnewline
\textbf{Hidden states} ($\times10^{3}$) & 193 & 63 & 200 & 71 & 209 & 81 & 471 & 153\tabularnewline
\bottomrule
\end{tabular}}
\end{table}

We also performed experiments with DenseNets; please see Table \ref{densenet results}.

\begin{table}
\caption{Classification error, computational requirements (in FLOPs), and memory load (measured by the number of hidden states) of two DenseNet networks with SVC convolutions on the CASIA dataset. The network architecture is specified in the top part of the table; the bottom part presents the performance characteristics of the networks. Lower is better. \label{densenet results}}
\vspace{3pt}
\begin{centering}
\resizebox{\linewidth}{!}{
\begin{tabular}{l|c|c}
\toprule
\multicolumn{1}{c|}{\textbf{Feature size}} & \textbf{DenseNet-A} & \textbf{DenseNet-B}\tabularnewline
\midrule
\multicolumn{1}{c|}{64$\times$64} & VSC(3,16,3,1), MP & VSC(3,16,3,1), MP\tabularnewline
\multicolumn{1}{c|}{32$\times$32} & $2\times16$ BC layers, Transition & $4\times16$ BC layers, Transition(Compress=0.5)\tabularnewline
\multicolumn{1}{c|}{16$\times$16} & $2\times16$ BC layers, Transition & $4\times16$ BC layers, Transition(Compress=0.5)\tabularnewline
\multicolumn{1}{c|}{8$\times$8} & $2\times16$ BC layers, Transition & 4$\times16$ BC layers, Transition(Compress=0.5)\tabularnewline
\multicolumn{1}{c|}{4$\times$4} & 2$\times16$ BC layers & $4\times16$ BC layers\tabularnewline
\multicolumn{1}{c|}{4$\times$4} & SC(144,256,4,1) & SC(128,256,4,1)\tabularnewline
\toprule
\textbf{Class. error} (in \%) & 4.49 & 4.09\tabularnewline
\textbf{FLOPs} ($\times10^{6}$)& 8.7 & 15\tabularnewline
\textbf{Hidden states} ($\times10^{3}$)  & 84 & 136\tabularnewline
\bottomrule
\end{tabular}}
\par\end{centering}
\end{table}

\begin{figure}[t]
\begin{centering}
\includegraphics[scale=0.6]{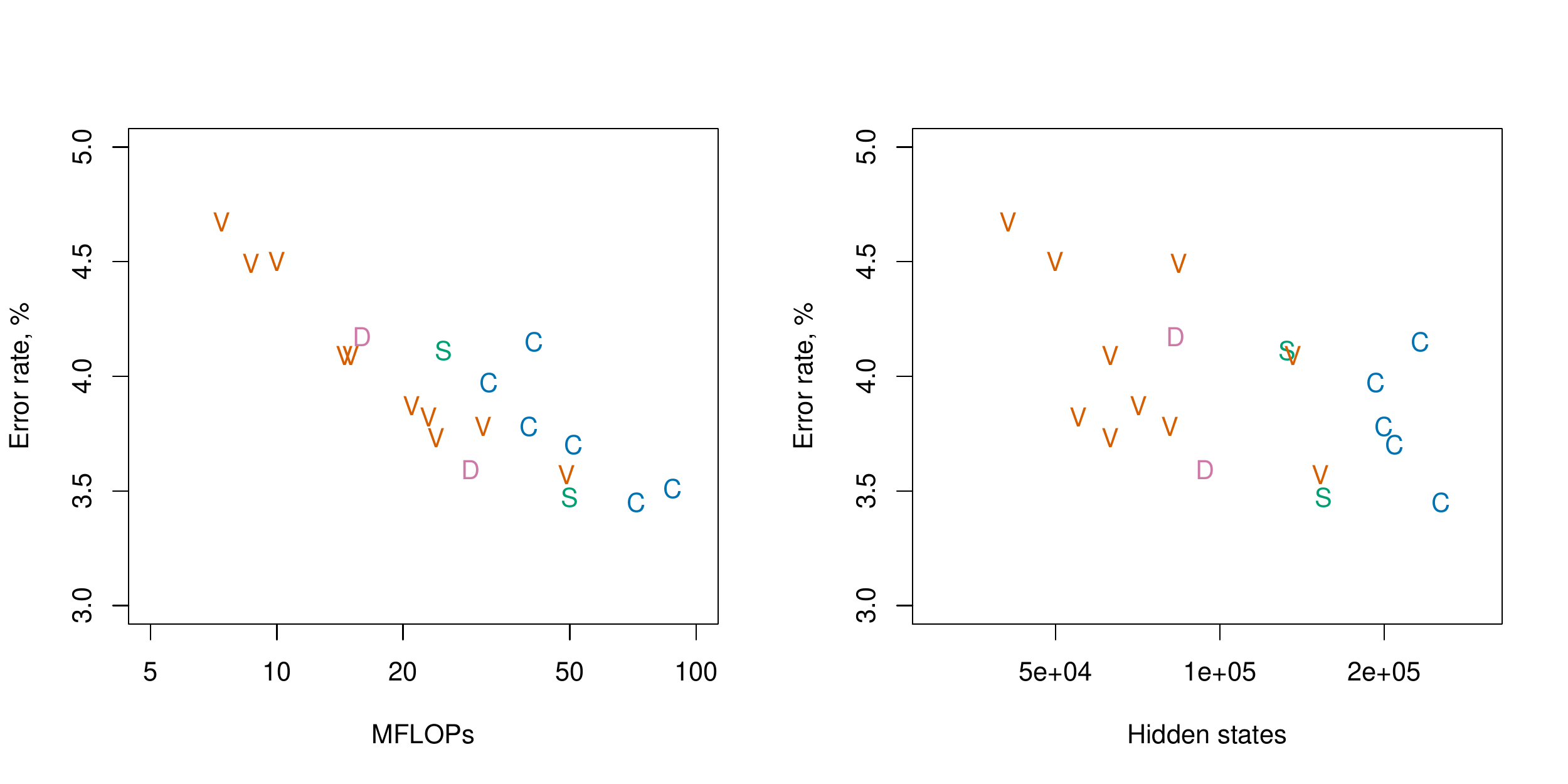}
\vspace{-4mm}
\caption{Overview of all experiments on the CASIA dataset. We denote submanifold convolutional networks by \textbf{V}, submanifold networks with SC-VSC-DC connections by \textbf{D}, regular sparse networks by \textbf{S}, and regular dense networks by \textbf{C}. Locations closer to the bottom-left corner are better.\label{KL}}
\par\end{centering}
\end{figure}

Next we experimented with adding extra connections to VGG networks to increase the effective receptive fields of the hidden states; see Table~\ref{vggCD} for results.
In the table, $\{a,b\}$ denotes a VSC$(\cdot,a,3,1)$ convolution performed in parallel with a chain of SC$(\cdot,b,3,2)$-VSC$(b,b,3,1)$-DC$(b,b,3,2)$ operations; outputs are concatenated to produce $a+b$ output feature planes. To simplify the network design, we switched to size-3 stride-2 max-pooling, matching the SC convolutions in the SC-VSC-DC branches, and reduce the input size from 64$\times$64 to 63$\times$63. Figure \ref{KL} presents an overview of all our results on the CASIA dataset.
\begin{table}
\begin{centering}
\caption{Evaluation of the effect of adding SC-VSC-DC branches to a VGG network on the CASIA dataset. Lower values are better. \label{vggCD}}
\vspace{3pt}
\par\end{centering}
\centering{}
{
\begin{tabular}{l|c|c|c|c}
\toprule
\multicolumn{1}{c|}{\textbf{Feature size}} & \textbf{VGG-C} & \textbf{VGG$^+$-C} & \textbf{VGG-D} & \textbf{VGG$^+$-D}\tabularnewline
\midrule
\multicolumn{1}{c|}{63$\times$63}  & $\{16,0\}\times2$  & $\{16,8\}\times2$ & $\{16,0\}\times2$  & $\{16,8\}\times2$\tabularnewline
\multicolumn{1}{c|}{31$\times$ 31} & $\{32,0\}\times2$  & $\{32,8\}\times2$ & $\{32,0\}\times2$  & $\{32,8\}\times2$\tabularnewline
\multicolumn{1}{c|}{15 $\times$15} & $\{48,0\}\times2$  & $\{48,16\}\times2$& $\{64,0\}\times2$  & $\{64,16\}\times2$\tabularnewline
\multicolumn{1}{c|}{7$\times$7}    & $\{64,0\}\times2$  & $\{64,16\}\times2$ & $\{128,0\}\times2$ & $\{96,16\}\times2$\tabularnewline
\multicolumn{1}{c|}{3$\times$3}    & $\{96,0\}\times2$  & $\{96,16\}\times2$ & $\{256,0\}\times2$ & $\{256,32\}\times2$\tabularnewline
\multicolumn{1}{c|}{3$\times$3}    & 128C3 & 128C3 & 512C3 & 512C3\tabularnewline
\midrule
\textbf{Class. error} (in \%)           & 4.50 & 4.17 & 3.73 & 3.59 \tabularnewline
\textbf{FLOPs} ($\times10^{6}$)         & 10   & 16 & 24   & 29\tabularnewline
\textbf{Hidden states} ($\times10^{3}$) & 50   & 83 & 63   & 94\tabularnewline
\bottomrule
\end{tabular}
}
\end{table}

\subsection{Results on ModelNet}
In a second set of experiments, we compare two submanifold VGG networks with a state-of-the-art dense convolutional network on the ModelNet-40 dataset \cite{oai:arXiv.org:1604.03265}. The results of these experiments are shown in Table \ref{vggM}: the left part of the table shows the architecture and performance of our submanifold VGG networks, whereas the right part of the table shows that of the dense 3DNiN network \cite{oai:arXiv.org:1604.03265}. The results clearly demonstrate that submanifold have the potential for designing convolutional networks for sparse data that obtain state-of-the-art performance with limited computational requirements: in particular, our VGG-A network makes 2\% more errors at 13$\times$ fewer computations, and our VGG-B performs roughly on par with the dense 3DNiN whilst performing $\sim\! 5 \times$ fewer computations.

\begin{table}[t]
\begin{centering}
\caption{Comparison of two submanifold VGG networks (left) with the dense 3DNiN network (right) on the ModelNet-40 dataset. Lower values are better.
\label{vggM}}
\vspace{3pt}
\par\end{centering}
\centering{}
\resizebox{\linewidth}{!}{
\begin{tabular}{l|c|c||c|c}
\toprule
\multicolumn{1}{c|}{\textbf{Feature size}} & \textbf{VGG-A} & \textbf{VGG-B} &\textbf{Feature size}& \textbf{3DNiN}\tabularnewline
\midrule
\multicolumn{1}{c|}{32$\times$ 32} & 3$\times$8C3, MP2 & 2$\times$16C3, MP2 &30$\times$30&48C6/2\tabularnewline
\multicolumn{1}{c|}{16 $\times$16} & 3$\times$16C3, MP2 & 2$\times$32C3, MP2 &13$\times$13&2$\times$48C1,96C5/4\tabularnewline
\multicolumn{1}{c|}{8 $\times$8} & 3$\times$24C3, MP2 & 2$\times$64C3, MP2 &5$\times$5&2$\times$96C1, 512C3/2\tabularnewline
\multicolumn{1}{c|}{4$\times$4} & 3$\times$32C3 & 2$\times$128C3 &2$\times$2&MP2\tabularnewline
\multicolumn{1}{c|}{4$\times$4} & 32C4 & 128C4 &--&--\tabularnewline
\toprule
\textbf{Convolution} & \textbf{VSC} & \textbf{VSC} & & \textbf{C}\tabularnewline
\midrule
\textbf{Class. error} (in \%) & 13.6 & 11.8 & \cellcolor{gray!50} &11.2\tabularnewline
\textbf{FLOPs} ($\times10^{6}$)& 9.2 & 28& \cellcolor{gray!50} &124\tabularnewline
\textbf{Hidden states} ($\times10^{3}$) & 55 & 78& \cellcolor{gray!50} &368\tabularnewline
\bottomrule
\end{tabular}
}
\end{table}

\section{Related Work}

This paper is not the first to study sparse convolutional networks \cite{Vote3Deep,Sparse3D,OctNet}. Most prior networks for sparse data implements a standard convolutional operator that increases the number of active sites with each layer \cite{Vote3Deep,Sparse3D}. By contrast, our submanifold convolutional networks allows sparse
data to be processed whilst retraining a much greater degree of sparsity. We have shown that this makes it practical to train deep and efficient VGG and ResNet models.

Submanifold convolutional networks are also much sparser than OctNets \cite{OctNet}.
OctNet stores data in oct-trees: a data structure in which the grid cube is progressively subdivided into
$2^3$ smaller sub-cubes until the sub-cubes are either empty or contain a single active site.
To compare the efficiency of OctNets with that of submanifold convolutional networks, we picked a random sample from the ModelNet-40 dataset and rendered it in a cube with $32^3$ grid points.
The resulting grid had 423 active sites, which corresponds to 1.3\% of the total number of sites.
Each active site had on average 12.4 active neighbors (the maximum possible number of neighbors is 27).
VSC convolutions, therefore, require only 0.6\% of the work of a dense (C) convolution.
However, in the OctTree, 80\%, 13\%, 4\%, and 3\% of the volume of the cube is covered by
sub-cubes of size $8^{3}$, $4^{3}$, $2^{3}$ and $1^{3}$, respectively.
As a result, an OctNet convolution, which operates over the surfaces of the smaller cubes,
requires about 35\% of the computations that a dense (C) convolution requires. In this particular example,  an OctNet convolution thus has a computational cost that is 60 times higher than that of a VSC convolution.

Submanifold convolutional networks also have advantages in terms of memory requirements. In particular, a submanifold network stores a single feature vector for each of the active sites. By contrast, OctTrees have about twice as many empty child nodes as active nodes, which implies they have to store roughly three times as many features as a submanifold convolutional network.

Having said that, some of the ideas of \cite{OctNet} may be combined with VSC convolutions. In particular, it is possible to use oct-trees as a specialized hash function in VSC convolutions. Such an oct-tree-based hash function has the potential to be faster than a standard universal hash function that operates on integer tuple keys, like in our implementation of VSC.

\section{Conclusion}

We introduced a new sparse convolutional operator, called \emph{valid sparse convolution} (VSC), that facilitates the design
of efficient, deep convolutional networks for sparse data. We have shown that
VSC convolutions lead to substantial computational savings whilst maintain state-of-the-art accuracies on two datasets: a dataset comprising one-dimensional manifolds embedded in two-dimensional space, and a dataset comprising two-dimensional surfaces embedded in three-dimensional space.

As part of this paper, we are releasing easy-to-use implementations of VSC and the other sparse operations we used in the networks described in this paper. We will also release code to reproduce the results of our experiments.

\bibliographystyle{abbrv}

\end{document}